\def\year{2020}\relax
\documentclass[letterpaper]{article} 
\usepackage{aaai20}  
\usepackage{times}  
\usepackage{helvet} 
\usepackage{courier}  
\usepackage[hyphens]{url}  
\usepackage{graphicx} 
\usepackage{graphicx}  
\usepackage{tabularx}
\usepackage{amsmath}
\usepackage{booktabs}
\usepackage{url}
\title{Context-Dependent Models for \\ Predicting and Characterizing Facial Expressiveness}
\author{Victoria Lin, Jeffrey M. Girard, Louis-Philippe Morency \\
Carnegie Mellon University \\
\{vlin2, jgirard2\}@andrew.cmu.edu, morency@cs.cmu.edu}
 
\begin{document}

\maketitle

\begin{abstract}
In recent years, extensive research has emerged in affective computing on topics like automatic emotion recognition and determining the signals that characterize individual emotions. Much less studied, however, is \emph{expressiveness}---the extent to which someone shows any feeling or emotion. Expressiveness is related to personality and mental health and plays a crucial role in social interaction. As such, the ability to automatically detect or predict expressiveness can facilitate significant advancements in areas ranging from psychiatric care to artificial social intelligence. Motivated by these potential applications, we present an extension of the BP4D+ dataset \cite{zhang2016multimodal} with human ratings of expressiveness and develop methods for (1) automatically predicting expressiveness from visual data and (2) defining relationships between interpretable visual signals and expressiveness. In addition, we study the emotional context in which expressiveness occurs and hypothesize that different sets of signals are indicative of expressiveness in different contexts (e.g., in response to surprise or in response to pain). Analysis of our statistical models confirms our hypothesis. Consequently, by looking at expressiveness separately in distinct emotional contexts, our predictive models show significant improvements over baselines and achieve comparable results to human performance in terms of correlation with the ground truth.
\end{abstract}

\section{Introduction}

Although humans constantly experience internal reactions to the stimuli around them, they do not always externally display or communicate those reactions. We refer to the degree to which a person does show his or her thoughts, feelings, or responses at a given point in time as \emph{expressiveness}. That is, a person being highly expressive at a given moment can be said to be passionate or even dramatic, whereas a person being low in expressiveness can be said to be stoic or impassive. In addition to varying moment-to-moment, a person's tendency toward high or low expressiveness in general can also be considered a trait or disposition \cite{fleeson2001toward}. 

In this paper, we study \textit{momentary} expressiveness, or expressiveness at a given moment in time. This quantity has not been previously explored in detail. We have two primary goals: (1) to automatically predict momentary expressiveness from visual data and (2) to learn and understand interpretable signals of expressiveness and how they vary in different emotional contexts. In the following subsections, we motivate the need for research on these two topics.

\subsubsection{Prediction of Expressiveness} The ability to automatically sense and predict a person's expressiveness is important for applications in artificial social intelligence and especially healthcare. For an example of how expressiveness might be useful in artifical social intelligence, as many customer-facing areas become increasingly automated, the computers, robots, and virtual agents who now interact with humans must be aware of expressiveness in order to interact with humans in appropriate ways (e.g., a highly expressive display might need to be afforded more attention than a less expressive one). With regard to healthcare, expressiveness holds promise as an indicator of mental health conditions like depression, mania, and schizophrenia, which have all been linked to distinct changes in expressiveness. Depression is associated with reduced expressiveness of positive emotions and increased expressiveness of certain negative emotions \cite{girardNonverbalSocialWithdrawal2014}; mania is associated with increased overall expressiveness \cite{nimhbipolar}; and schizophrenia is associated with blunted expressiveness and inappropriate affect, or expressiveness for the ``wrong'' emotion given the context \cite{hamm2011automated}. Because these relationships are known, predicting an individual's expressiveness can provide a supplemental measure of the presence or severity of specific mental health conditions. An automatic predictor of expressiveness therefore has the potential to support clinical diagnosis and assessment of behavioral symptoms.

\subsubsection{Understanding Signals of Expressiveness} Intuitively, overall impressions of expressiveness are grounded in visual signals like facial expression, gestures, body posture, and motion. However, the signals that correspond to high expressiveness in a particular emotional context do not necessarily correspond to high expressiveness in a different emotional context. For example, a person who has just been startled may express his or her reaction strongly by flinching, which results in a fast and large amount of body movement. On the other hand, a person who is in pain may show that feeling by moving slowly and minimally because he or she is attempting to regulate their emotion. In the former scenario, quick movement corresponds to high expressiveness, whereas in the latter scenario, quick movement corresponds to low expressiveness.

We aim to formalize the relationship between interpretable visual signals and expressiveness through statistical analysis. Furthermore, we hypothesize that the specific signals that contribute to expressiveness vary somewhat under different contexts and seek to confirm this hypothesis by modeling expressiveness in different emotional states.

\subsubsection{Contributions} To realize our goals, we must collect data about how expressiveness is perceived in spontaneous (i.e., not acted) behavior and develop techniques to analyze, model, and predict it. As such, we address the gap in the literature through the following contributions.
\begin{itemize}
    \item We introduce an extension of the BP4D+ emotion elicitation dataset \cite{zhang2016multimodal} with human ratings of central aspects of expressiveness: response strength, emotion intensity, and body and facial movement. We also describe a method for generating a single expressiveness score from these ratings using a latent variable representation of expressiveness.
    \item We present statistical and deep learning models that are able to predict expressiveness from visual data. We perform experiments on a test set of the BP4D+ extended dataset, establish baselines, and show that our models are able to significantly outperform those baselines and for some metrics even approach human performance, particularly when taking context into consideration.
    \item We present context-specific and context-agnostic statistical models that reveal interpretable relationships between visual signals and expressiveness. We conduct an analysis of these relationships over three emotional contexts---startle, pain, and disgust---that supports our hypothesis that the set of visual signals that are important to expressiveness varies depending on the emotional context.
\end{itemize}

\section{Related Work}

Although little prior work has been conducted on direct prediction of expressiveness, advances have been made in the adjacent field of emotion recognition. Likewise, within the scope of psychology, there exists a substantial body of literature dedicated to determining the visual features that characterize different emotions; however, to our knowledge, little to no similar work has been conducted on the visual features that characterize how strongly those emotions are shown (i.e., expressiveness). We describe the current state of these areas of research, as we draw from this related work to define our own approaches to predicting and characterizing expressiveness. 

\subsubsection{Emotion Recognition}
Because the task derives from similar visual features---facial landmarks and movement, for example---advancements in deep learning for the field of emotion recognition are highly informative and provide much of the guiding direction for our predictive deep learning models. A number of architectures have achieved high accuracy for multiclass emotion classification in a variety of settings, including still images, videos, and small datasets. \cite{yu2015image} used an ensemble of CNNs with either log-likelihood or hinge loss to classify images of faces from movie stills as belonging to 1 of 7 basic emotions. \cite{ng2015deep} extended a similar architecture to accurately predict emotions even with little task-specific training data by performing sequential fine-tuning of a CNN pretrained on ImageNet, first with a facial expression dataset and then with the target dataset, a small movie still dataset. \cite{byeon2014facial} designed a 3D-CNN that predicts the presence of an emotion (as opposed to a neutral expression) in each frame of a video. Finally, \cite{ebrahimi2015recurrent} proposed a hybrid approach for emotion recognition in video. After first training a CNN on two separate datasets of static images of facial emotions, the authors used the CNN to obtain embeddings of each frame, which they used as sequential inputs to an RNN to classify emotion.

\subsubsection{Interpretable Signals of Emotion}
The three emotional contexts of startle, pain, and disgust all have well-studied behavioral responses that could serve as visual signals of emotion and therefore expressiveness. Previous observational research has found that the human startle response is characterized by blinking, hunching the shoulders, pushing the head forward, grimacing, baring the teeth, raising the arms, tightening the abdomen, and bending the knees \cite{sillarMammalianStartleResponse2016}; the human pain response is characterized by facial grimacing, frowning, wincing, increased muscle tension, increased body movement/agitation, and eye closure \cite{kunzFacialMuscleMovements2019}; and the human disgust response is characterized by furrowed eyebrows, eye closure, pupil constriction, nose wrinkling, upper lip retraction, upward movement of the lower lip and chin, and drawing the corners of the mouth down and back \cite{tyburDisgustEvolvedFunction2013,olatunjiDisgustCharacteristicFeatures2005}. These responses have notable similarities, such as the presence of grimacing, eye closure, and withdrawal from an unpleasant stimulus. However, they also have unique aspects, such as pushing the head forward in startle, increased muscle tension in pain, and nose wrinkling in disgust.

\section{Expressiveness Dataset}
We describe the data collection pipeline and engineering process for the dataset we used to perform our modeling and analysis of expressiveness.

\subsubsection{Video Data}

\begin{figure}
    \centering
    \includegraphics[width=\linewidth]{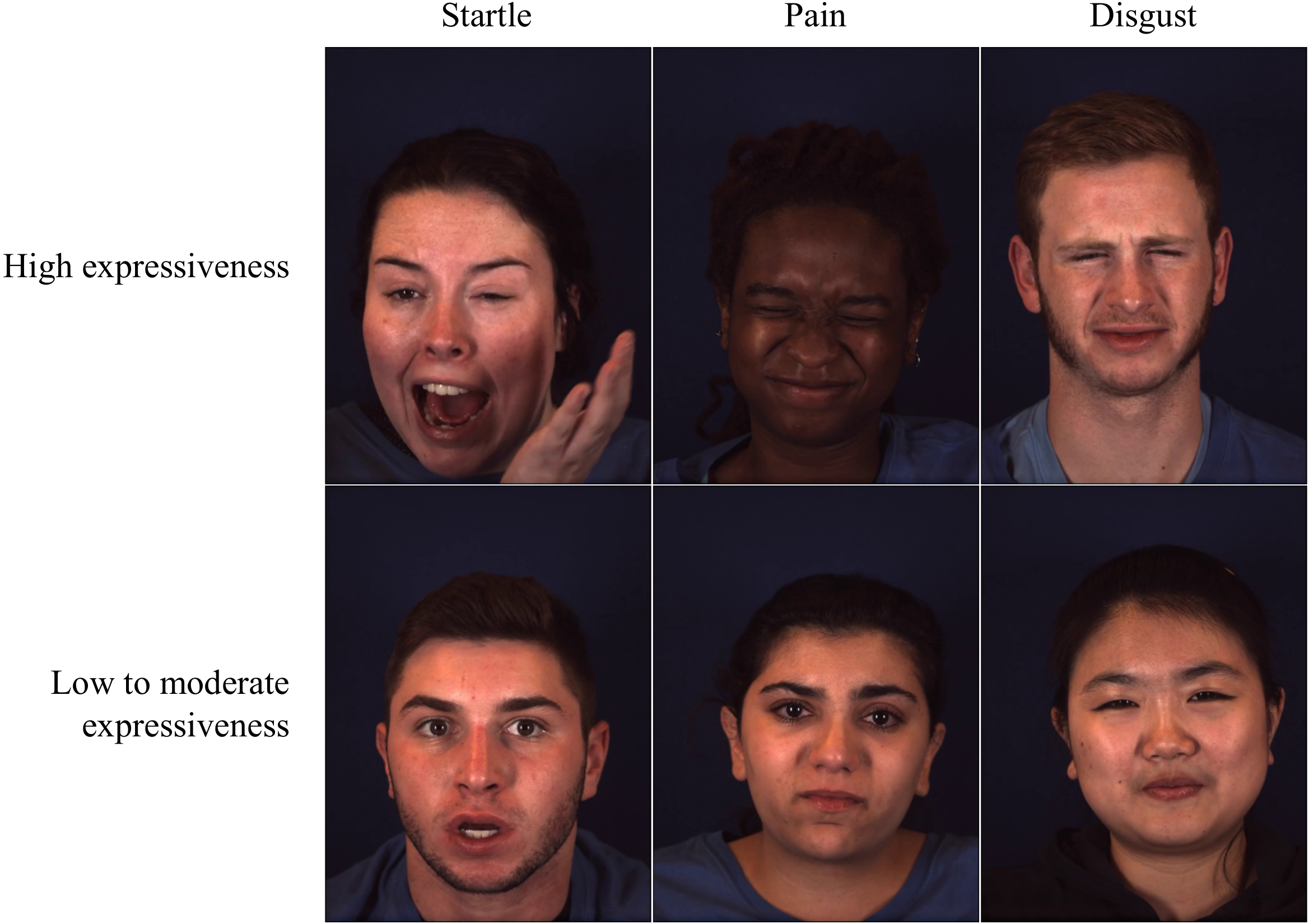}
    \caption{Example frames from videos of different emotion elicitation tasks in the BP4D+ dataset.}
    \label{fig:data_samples}
\end{figure}

The BP4D+ dataset contains video and metadata of 140 participants performing ten tasks meant to elicit ten different emotional states \cite{zhang2016multimodal}. Participants were mostly college-aged ($M=21.0$, $SD=4.9$) and included a mix of genders and ethnicities ($59\%$ female, $41\%$ male; $46\%$ White, $33\%$ Asian, $11\%$ Black, $10\%$ Latinx). A camera captured high definition images of participants' faces during each task at a rate of 25 frames per second. On average, tasks lasted $44.5$ seconds in duration ($SD=31.4$).

In this study, we focus on the tasks meant to elicit startle, pain, and disgust. Example frames from each of these tasks can be found in Figure \ref{fig:data_samples}. These tasks were selected because they did not involve the participant talking; we wanted to avoid tasks involving talking because the audio recordings are not available as part of the released dataset. In the startle task, participants unexpectedly heard a loud noise behind them; in the pain task, participants submerged their hands in ice water for as long as possible; and in the disgust task, participants smelled an unpleasant odor similar to rotten eggs. 

Because a person's expressiveness may change moment-to-moment and we wanted to have a fine-grained analysis, we segmented each task video into multiple 3\nobreakdash-second clips. Because task duration varied between tasks and participants, and we did not want examples with longer durations to dominate those with shorter durations, we decided to focus on a standardized subset of video clips from each task. For the startle task, we focused on the five clips ranging from second 3 to second 18 as this range would capture time before, during, and after the loud noise. For the pain task, we focused on the first three clips when pain was relatively low and the final four clips when pain was relatively high. Finally, for the disgust task, we focused on the four clips ranging from second 3 to second 15 as this range would capture time before, during, and after the unpleasant odor was introduced. In a few cases, missing or dropped video frames were replaced with empty black images to ensure a consistent length of 3 seconds per clip.

\subsubsection{Human Annotation}
We defined expressiveness as the degree to which others would perceive a person to be feeling and expressing emotion. Thus, we needed to have human annotators watch each video clip and judge how expressive the person in it appeared to be. To accomplish this goal, we recruited six crowdworkers from Amazon's Mechanical Turk platform to watch and rate each video clip. We required that raters be based in the United States and have approval ratings of $99\%$ or greater on all previous tasks. Raters were compensated at a rate approximately equal to $\$7.25$ per hour.

Because raters may have different understandings of the word ``expressiveness,'' we did not want to simply ask them to rate how expressive each clip was. Instead, we generated three questions intended to directly capture important aspects of expressiveness. Specifically, we asked: (1) How strong is the emotional response of the person in this video clip to [the stimulus] compared to how strongly a typical person would respond? (2) How much of any emotion does the person show in this video clip? (3) How much does the person move any part of their body/head/face in this video clip? Each question was answered using a five-point ordered scale from 0 to 4 (see the appendix for details).

To assess the inter-rater reliability of the ratings (i.e., their consistency across raters), we calculated intraclass correlation coefficients (ICC) for each question in each task and across all tasks. Because each video clip was rated by a potentially different group of raters, and we ultimately analyzed the average of all raters' responses (as described in the next subsection), the appropriate ICC formulation is the one-way average score model \cite{mcgraw1996forming}. ICC coefficients at or above 0.75 are often considered evidence of ``excellent'' inter-rater reliability \cite{cicchetti1994guidelines}. As shown in Table \ref{tab:turk-icc}, all the ICC estimates---and even the lower bounds of their 95\% confidence intervals---exceeded this threshold. Thus, inter-rater reliability was excellent.

\begin{table}
    \begin{center}
    
    \begin{tabular}{llcc}
        \toprule
        Task & Question & ICC & \multicolumn{1}{c}{95\% CI} \\
        \midrule
        Startle & 1~~(Response) & 0.84 & [0.82, 0.86] \\
        Startle & 2~~(Emotion) & 0.85 & [0.83, 0.87] \\
        Startle & 3~~(Motion) & 0.85 & [0.84, 0.87] \\
        \midrule
        Pain & 1~~(Response) & 0.84 & [0.82, 0.85] \\
        Pain & 2~~(Emotion) & 0.83 & [0.81, 0.85] \\
        Pain & 3~~(Motion) & 0.80 & [0.78, 0.82] \\
        \midrule
        Disgust & 1~~(Response) & 0.88 & [0.87, 0.90] \\
        Disgust & 2~~(Emotion) & 0.88 & [0.87, 0.90] \\
        Disgust & 3~~(Motion) & 0.86 & [0.84, 0.88] \\
        \midrule
        All & 1~~(Response) & 0.86 & [0.85, 0.87] \\
        All & 2~~(Emotion) & 0.86 & [0.85, 0.87] \\
        All & 3~~(Motion) & 0.85 & [0.84, 0.86] \\
        \bottomrule
    \end{tabular}
    \end{center}
\caption{\label{tab:turk-icc}Inter-rater reliability of crowdworkers per question.}
\end{table}

\subsubsection{Expressiveness Scores} For each video clip, we wanted to summarize the answers to each of the three questions asked as a single expressiveness score to use as our target in machine learning and statistical analysis, as each question captured an important aspect of expressiveness. Each of the six raters assigned to each video clip provided three answers. The simplest approach to aggregating these 18 scores would be to average them. However, this would assume that all three questions are equally important to our construct of expressiveness and equally well-measured. To avoid this assumption, we first calculated the average answer to each question across all six raters and then used confirmatory factor analysis (CFA) to estimate a latent variable that explains the variance shared amongst the questions \cite{klinePrinciplesPracticeStructural2015}. 

\begin{figure}[]
    \includegraphics[width=.5\linewidth]{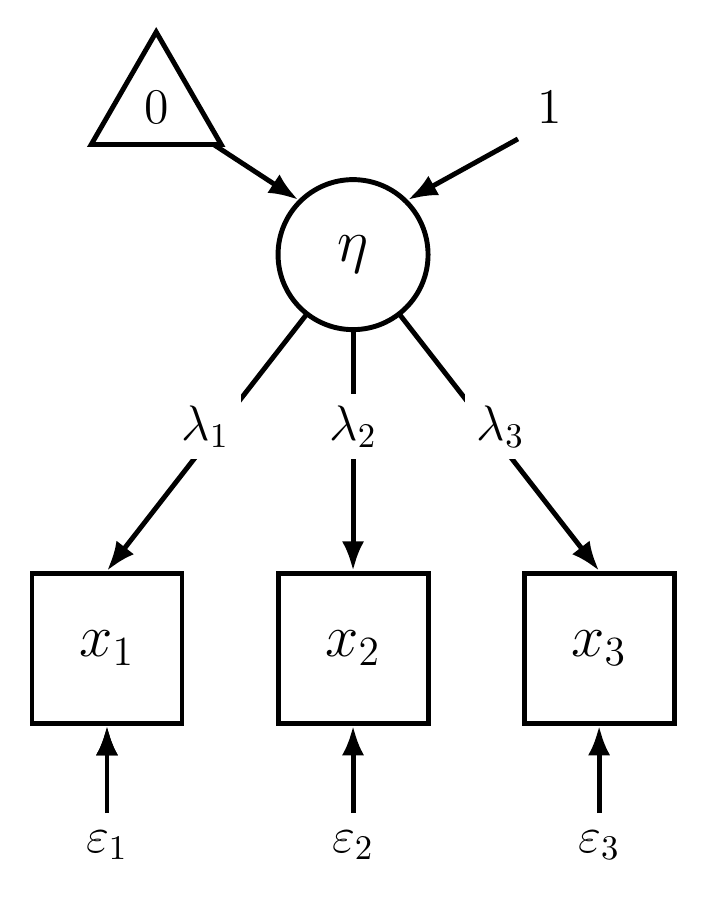}
    \centering
    \caption{Diagram of confirmatory factor analysis.}
    \label{fig:cfa}
\end{figure}

In Figure~\ref{fig:cfa}, the observed question variables are depicted as squares $(x)$ and the aforementioned latent variable is depicted as a circle $(\eta)$ with zero mean and unit variance. The factor loadings $(\lambda)$ represent how much each question variable was composed of shared variance, and the residuals $(\varepsilon)$ represent how much each question variable was composed of non-shared variance (including measurement error). We fit this same CFA model for each task separately and across all tasks using the \texttt{lavaan} package \cite{rosseelLavaanPackageStructural2012}. 

The resulting estimates are provided in Table~\ref{tab:loadings}. Three patterns in the results are notable. First, all the standardized loadings were higher than 0.85 (and most were higher than 0.95), which suggests that there is a great deal of shared variance between these questions and they are all measuring the same thing (e.g., expressiveness). Second, there were some factor loading differences within tasks, which suggests that there is value in aggregating the question responses using CFA rather than averaging them. Third, there were some factor loading differences between tasks, especially for the motion question, which suggests that the relationship between motion and expressiveness depends upon context.

Finally, we estimated each video clip's standing on the latent variable (i.e., as a continuous, real-valued number) by extracting factor score estimates from the CFA model; this was done using the Bartlett method, which produces unbiased estimates of the true factor scores \cite{distefanoUnderstandingUsingFactor2009}. These estimates were then used as ground truth expressiveness labels in our further analyses.

\begin{table}[]
    \centering
    \begin{tabular}{l c c c c}
    \toprule
     & Startle & Pain & Disgust & All \\
    \midrule
    $\lambda_1$ (Response)      & 0.98 & 0.98 & 0.98 & 0.98 \\
    $\lambda_2$ (Emotion)       & 0.97 & 0.96 & 0.98 & 0.97 \\
    $\lambda_3$ (Motion)        & 0.95 & 0.85 & 0.90 & 0.91 \\
    \midrule
    $\varepsilon_1$ (Response)  & 0.05 & 0.04 & 0.04 & 0.04 \\
    $\varepsilon_2$ (Emotion)  & 0.07 & 0.07 & 0.04 & 0.06 \\
    $\varepsilon_3$ (Motion)  & 0.10 & 0.28 & 0.19 & 0.18 \\
    \bottomrule
    \end{tabular}
    \caption{Model parameter estimates from confirmatory factor analysis.}
    \label{tab:loadings}
\end{table}

\section{Methods}
We selected our models with our two primary goals in mind: we wanted to find a model that would perform well in predicting expressiveness, and we wanted at least one interpretable model so that we could understand the relationships between the behavioral signals and the expressiveness scores. We experimented with three primary architectures---ElasticNet, LSTM, and 3D-CNN---and describe our approaches in greater detail below.

\subsubsection{ElasticNet}
We chose ElasticNet \cite{zouRegularizationVariableSelection2005} as an approach because it is suitable for both our goals of prediction and interpretation. ElasticNet is essentially linear regression with regularization by a mixture of L1 and L2 priors. This regularization eliminates the problems of overfitting and multicollinearity common to linear regression with many features and achieves robust generalizability. However, ElasticNet is still fully interpretable: examination of the feature weights provides insight into the relationships between features and labels.

We engineered visual features from the raw video data to use as input for our ElasticNet model. For each clip, we used the OpenFace \cite{baltrusaitis2018openface} toolkit to extract per-frame descriptors of gaze, head pose, and facial landmarks (e.g., eyebrows, eyes, mouth), as well as estimates of the occurrence and intensity for a number of action units from the Facial Action Coding System \cite{ekmanFacialActionCoding2002}. To reduce the effects of jitter, which may produce differences from frame to frame due simply to noise, we downsampled our data to 5 Hz from the original 25 Hz. 

From this data, we computed frame-to-frame displacement (i.e., distance travelled) and velocity (i.e., the derivative of displacement) for each facial landmark. We also calculated frame-to-frame changes in gaze angle and head position with regard to translation and scale (``head''); pitch; yaw; and roll. For each clip, we used the averages over all frames of these quantities as our features. We also counted the total number of action units and calculated the mean intensity of action units occurring in the clip. We selected these features to represent both amount and speed of facial, head, and gaze movement.

We used an out-of-the-box implementation of ElasticNet from \texttt{sklearn} and tuned the hyperparameters by searching over $\alpha \in \{0.01, 0.05, 0.1, 0.5, 1.0\}$ for the penalty term and over $\lambda \in \{0.0, 0.1, \dots, 1.0\}$ for the L1 prior ratio. For the final models on the startle task, pain task, disgust task, and all tasks, $\alpha$ was $0.01$, $0.1$, $0.1$, and $0.05$, respectively; $\lambda$ was $0.0$, $0.0$, $0.7$, and $0.7$, respectively. When $\lambda=0.0$, ElasticNet corresponds to Ridge regression, and when $\lambda=1.0$, ElasticNet corresponds to Lasso regression \cite{zouRegularizationVariableSelection2005}.

\subsubsection{OpenFace-LSTM}
We also explored several deep learning approaches to determine whether we could achieve better predictive performance by sacrificing some interpretability. Due to the small size of the training dataset and the need to capture the temporal component of the data, we proposed the use of a relatively simple deep architecture suitable for modeling sequences of data, LSTM \cite{hochreiter1997long}. We implemented our LSTM from scratch using the PyTorch framework and tuned over learning rate, number of layers, and hidden dimension of each layer. In our final implementation, we used learning rate $0.005$ with $2$ layers of hidden dimension $128$. Rather than engineering summary features as we did for ElasticNet, we used a tensor representation of the raw OpenFace facial landmark point tracking descriptors for each clip as input for the LSTM. Because the LSTM is more capable of handling high-dimensional data than a linear model, we retained the original sample rate of 25 Hz to reduce loss of information. Each clip with 75 frames was represented as a [$75 \times 614$] 2-dimensional tensor, where we standardized each $[75 \times 1]$ feature by subtracting its mean and dividing by its standard deviation.

\subsubsection{3D-CNN}
Although manual feature engineering can be useful for directing models to use relevant visual characteristics to make their predictions, it can also result in the loss of large amounts of information and furthermore has the potential to introduce noise. Consequently, we also explored the predictive performance of deep learning models that learn their own feature representations from the raw video data. Drawing on past successes with similar architectures in the related topic of emotion recognition, we selected as our model 3D-CNN \cite{ji20123d}, which is also capable of handling the temporal aspect of our data. Our 3D-CNN predicts expressiveness directly from a video clip. We modified the 18-layer Resnet3D available through PyTorch's \texttt{torchvision} \cite{tran2018closer} to perform prediction of a continuous value rather than classification, while retaining the hyperparameter values of the original implementation. We experimented both with training the model from scratch on the BP4D+ extension dataset and with using the BP4D+ extension only for fine-tuning of a 3D-CNN pretrained on the Kinetics 400 action recognition dataset \cite{kay2017kinetics}.

\section{Experiments}
In this section, we describe the evaluation metrics, data partitions, and baselines that we used to evaluate the performance of our models and to conduct our analysis of the interpretable visual features relevant to expressiveness. Code for our evaluation and analyses is available at \url{https://osf.io/bp7df/?view_only=70e91114627742d7888fbdd36a314ee9}.

\subsubsection{Evaluation Metrics and Dataset}
We selected RMSE and correlation of model predictions with the ground truth expressiveness scores as the evaluation metrics for our model performance. For ease of interpretability and comparison, we report normalized RMSE \cite{luo2016guidelines}, which we define as the RMSE divided by the scale of the theoretical range of the expressiveness scores. The value of the normalized RMSE ranges from 0 to 1, with 0 being the best performance and 1 being the worst performance.

To determine whether differences in performance between models and baselines were statistically significant, we used the cluster bootstrap \cite{fieldBootstrappingClusteredData2007,renNonparametricBootstrappingHierarchical2010} to generate 95\% confidence intervals and $p$-values for the differences in RMSEs and correlations between models. This approach does not make parametric assumptions about the distribution of the difference scores and accounts for the hierarchical dependency of video clips within subjects.\footnote{Software to conduct this procedure is available at \url{https://github.com/jmgirard/mlboot}.}

\begin{table*}
    \begin{center}
    \begin{tabular}{lccccccccc}
        \toprule
        & \multicolumn{4}{c}{Normalized RMSE (lower is better)} & & \multicolumn{4}{c}{Correlation (higher is better)} \\
        \cmidrule{2-5}\cmidrule{7-10}
         & Startle & Pain & Disgust & All && Startle & Pain & Disgust & All \\
        \midrule
        Uniform baseline    & $0.294$ & $0.303$ & $0.323$ & $0.309$ & & $0.091$ & $0.032$ & $-0.078$ & $0.043$\\
        Normal baseline       & $0.211$ & $0.196$ & $0.198$ & $0.210$ & & $-0.039$ & $-0.039$ & $0.084$ & $0.023$\\
        Human baseline      & $0.087$ & $0.093$ & $0.072$ & $0.081$ & & $0.768$ & $0.698$ & $0.831$ & $0.792$ \\
        \midrule
        3D-CNN              & $0.156$ & $0.122$ & $0.150$ & $0.148$ & & $-0.141$ & $0.127$ & $0.169$ &  $0.015$ \\
        3D-CNN pretrained   & $0.152$ & $0.116$ & $0.149$ & $0.148$ & & $0.232$ & $0.338$ & $0.053$ & $0.129$ \\
        OpenFace-LSTM         & $0.129$ & $0.124$ & $0.127$ & $0.145$ & & $0.508$ & $0.031$ & $0.538$ & $0.276$ \\
        \midrule
        ElasticNet               & $\mathbf{0.100}$ & $\mathbf{0.104}$ & $\mathbf{0.084}$ & $\mathbf{0.124}$ & & $\mathbf{0.723}$ & $\mathbf{0.525}$ & $\mathbf{0.834}$ & $\mathbf{0.497}$ \\
        \bottomrule
    \end{tabular}
    \end{center}
\caption{\label{tab:model-rmses}Test performance by task and overall on predicting expressiveness.}
\end{table*}

\begin{figure}[t]
    \centering
    \includegraphics[width=\linewidth]{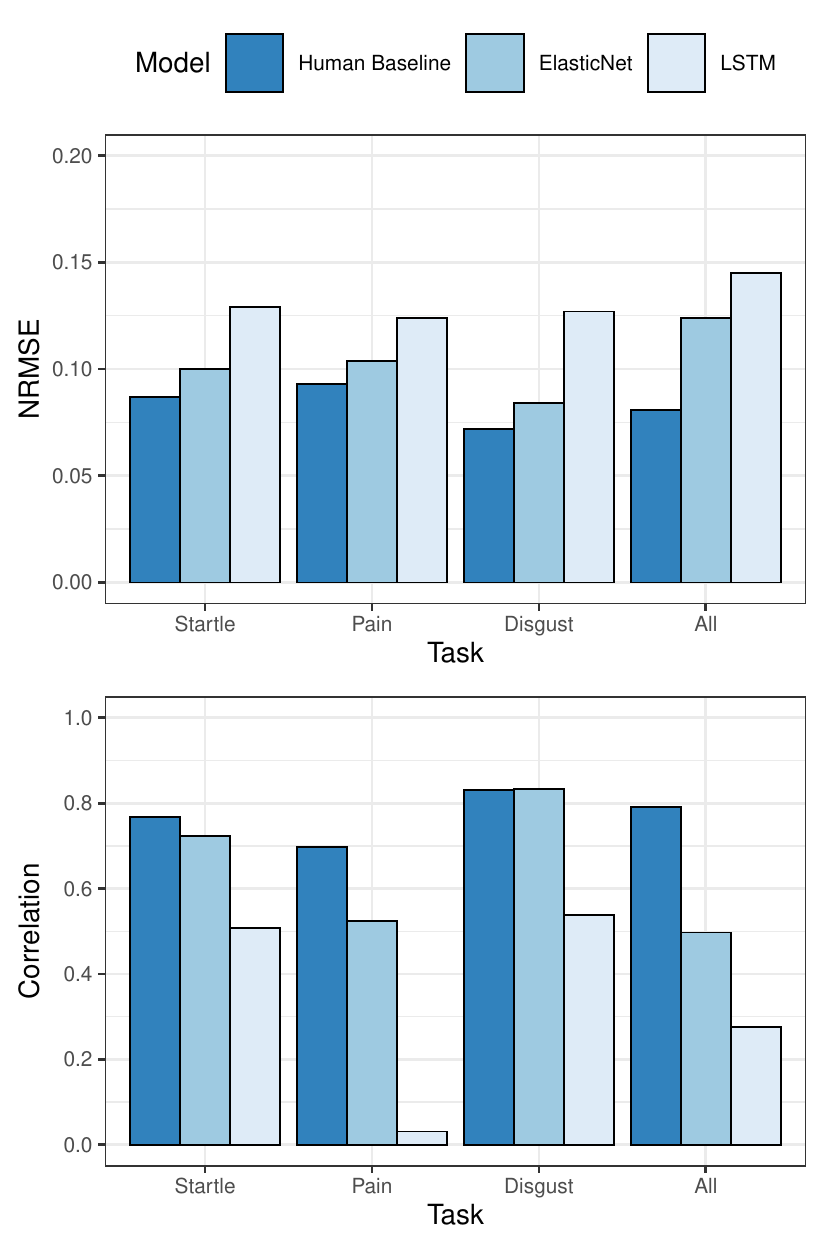}
    \caption{Performance comparison across models by task. Better performance is indicated by a lower NRMSE (range: $0$ to $1$) and a higher correlation (range: $-1$ to $1$).}
    \label{fig:bar}
\end{figure}

\begin{table*}
    \begin{center}
    \begin{tabular}{lccccccccc}
        \toprule
        & \multicolumn{3}{c}{$\Delta$ NRMSE over all tasks} & & \multicolumn{3}{c}{$\Delta$ Correlation over all tasks} \\
        \cmidrule{2-4}\cmidrule{6-8}
         & Estimate & 95\% CI & $p$-value & & Estimate & 95\% CI & $p$-value \\
        \midrule
        EN $-$ Uniform baseline & $-0.185$ & $[-0.203, -0.168]$ & $<0.001$ & & $0.389$ & $[0.218, 0.533]$ & $<0.001$ \\
        EN $-$ Normal baseline & $-0.086$ & $[-0.104, -0.071]$ & $<0.001$ & & $0.401$ & $[0.262, 0.523]$ & $<0.001$ \\
        EN $-$ Human baseline & $0.034$ & $[0.016, 0.052]$ & $0.001$ & & $-0.333$ & $[-0.456, -0.205]$ & $<0.001$ \\
        \midrule
        EN $-$ 3D-CNN   & $-0.021$ & $[-0.036, -0.011]$ & $0.001$ & & $0.413$ & $[0.209, 0.586]$ & $<0.001$\\
        EN $-$ 3D-CNN pretrained & $-0.022$ & $[-0.035, -0.012]$ & $<0.001$ & & $0.433$ & $[0.260, 0.560]$ & $<0.001$ \\
        EN $-$ OpenFace-LSTM & $-0.019$ & $[-0.034, -0.009]$ & $0.003$ & & $0.213$ & $[0.082, 0.343]$ & $<0.001$  \\
        \bottomrule
    \end{tabular}
    \end{center}
\caption{\label{tab:mlboot-results}Comparison of ElasticNet performance with performance of all other baselines and models. $\Delta \text{ NRMSE} < 0$ and $\Delta \text{ Corr} > 0$ indicate that ElasticNet performs better relative to the other model.}
\end{table*}

Because we suspected that expressiveness might manifest differently in different emotions, we wanted to see whether training separate models for each emotion elicitation task would produce better predictive performance than training a single model over all tasks. Furthermore, fitting separate ElasticNet models for each task would allow us to understand whether the feature set relevant to expressiveness is different depending on the emotional context, which would test our hypothesis. Therefore, we separated the BP4D+ dataset by task and created $60/20/20$ train/validation/test splits for each of these task-specific datasets and a separate split in the same proportions over the entire dataset. This partitioning was done such that no subject appeared in multiple splits. For each model, we report results from training and evaluating on each task-specific dataset and on the entire dataset.

\subsubsection{Baselines}
We defined several baselines against which to compare our models' performance:
\begin{itemize}
    \item \textbf{Uniform baseline}: This baseline samples randomly from a uniform distribution over the theoretical range of the expressiveness scores (i.e., $-3.5$ to $3.5$).
    \item \textbf{Normal baseline}: This baseline samples randomly from a standard normal distribution with mean and variance equal to the theoretical mean and variance of the expressiveness scores (i.e., mean $0$ and variance $1$).
    \item \textbf{Human baseline}: This baseline represents the performance of a single randomly selected human crowdworker. We calculated an estimated factor score for each rater by weighting their answers to each question by that question's factor loading and summing the weighted values. These weighted sums were then standardized and compared to 
    the average of the remaining 5 raters' estimated factor scores 
    to assess each rater's solitary performance. Finally, these performance scores were averaged over all crowdworkers to capture the performance of a randomly selected crowdworker.
\end{itemize}

\section{Results and Discussion}
In the following subsections, we present the results of our experiments, first comparing our model approaches and baselines and then visualizing and interpreting the feature weights of the ElasticNet model.

\begin{figure*}[t]
    \includegraphics[]{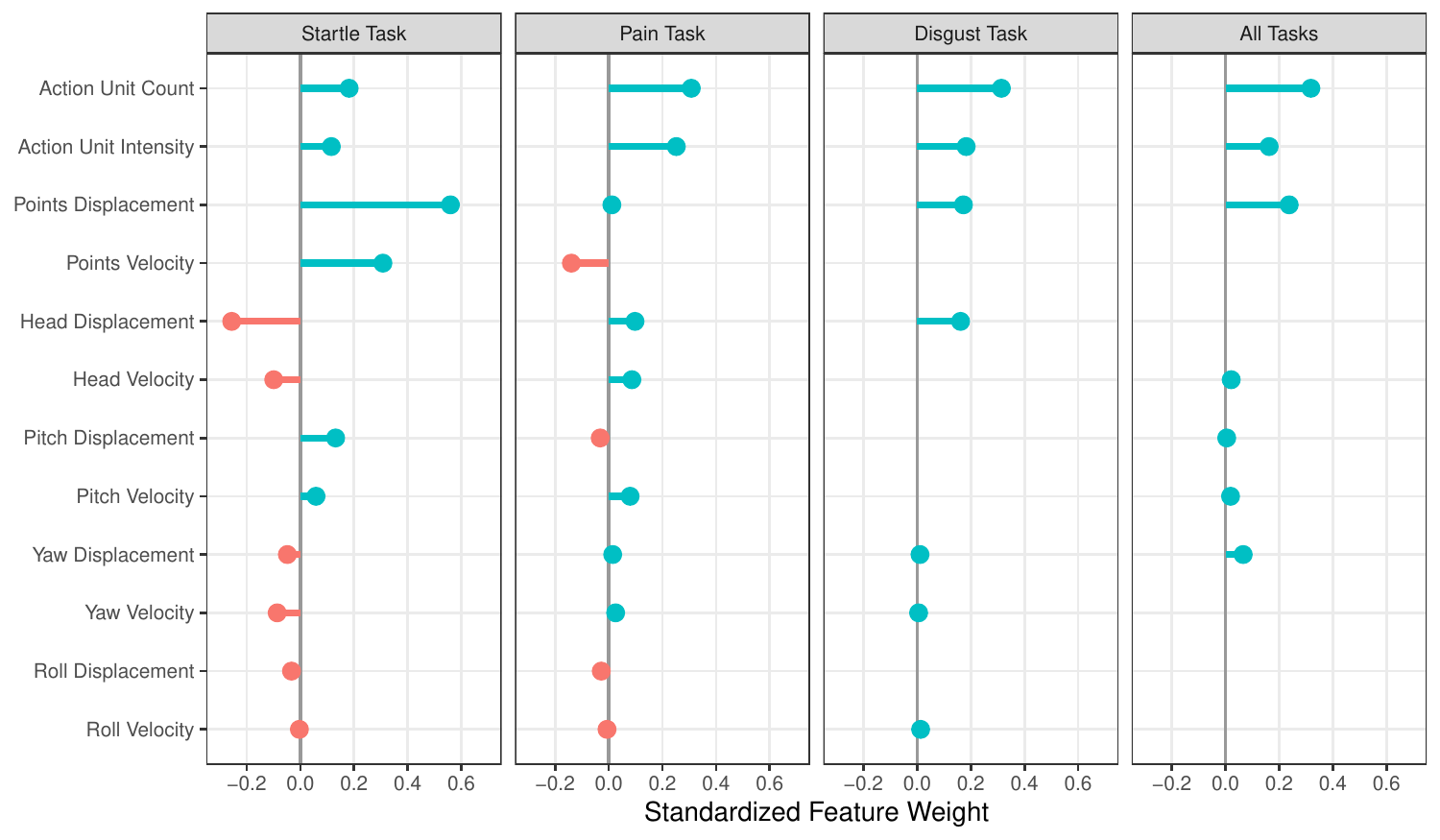}
    \centering
    \caption{Feature weights for each task-specific ElasticNet model, as well as for the model over all tasks.}
    \label{fig:weights}
\end{figure*}

\subsubsection{Prediction of Expressiveness}
The results of our performance evaluation are provided in Table~\ref{tab:model-rmses} and depicted in Figure~\ref{fig:bar}. Our three proposed approaches all show substantially improved performance over a simple method like the normal baseline. In particular, ElasticNet produced the lowest NRMSEs and highest correlations of the proposed methods on all individual tasks and over all tasks combined.  

Despite achieving NRMSEs well below those of the normal baseline, the proposed deep learning had relatively poor performance in most tasks according to the correlation metric. For example, OpenFace-LSTM attained a reasonable correlation compared to the human baseline on the startle and disgust tasks but produced essentially no correlation with the ground truth on the pain task. Likewise, pretrained 3D-CNN and 3D-CNN trained from scratch yielded little and no correlation, respectively, of their predictions with the ground truth. We suspect that such results may be the product of the small dataset on which the models were trained, as the data quantity may be insufficient to allow the models to generalize and learn the appropriate predictive signals from complex data without human intervention.

As such, of the proposed models, we consider ElasticNet to demonstrate the best performance overall. Its NRMSEs were consistently lower than those of the other proposed models, and its correlations were much higher than those of any other proposed model and come close to (and in the case of the disgust task, slightly exceed) those of the human baseline. Statistical analyses of the differences in performance between ElasticNet and all other models and baselines, the results of which are shown in Table~\ref{tab:mlboot-results}, support our intuition. Specifically, when trained across all tasks, ElasticNet attains significantly lower NRMSE and significantly higher correlation of its predictions with the ground truth compared to all other models and baselines except the human baseline. However, the same comparison also shows that ElasticNet has significantly higher NRMSE and significantly lower correlation of its predictions with the ground truth compared to the human baseline, indicating that there is still room for improvement.

\subsubsection{Understanding Signals of Expressiveness}

Because our best-performing model, ElasticNet, is an interpretable linear model, we were able to determine the relationship between the visual features in our dataset and overall expressiveness by examining the feature weights of the model trained over all tasks. Furthermore, by doing the same for the feature weights of models trained over individual tasks, we were able to explore the hypothesis that the set of signals indicative of expressiveness varies from context to context. These visualizations are shown in Figure~\ref{fig:weights}. We directly interpret those features with a standardized weight close to or greater than $0.2$ in absolute value.

From the weights of the model trained over all tasks, we can see that three primary features contribute to predicting overall expressiveness: action unit count, action unit intensity, and point displacement (i.e., the distance traveled by all facial landmark points). This suggests that there are some behavioral signals that index expressiveness across emotional contexts, and these are generally related to the amount and intensity of facial motion. Notably, features related to head motion and the velocity of motion did not have high feature weights for overall expressiveness.

We also observe that each individual task had its own unique set of features that were important to predicting expressiveness within that context. These features make intuitive sense when considering the nature of the tasks and are consistent with the psychological literature we reviewed. 

In the startle task, higher expressiveness was associated with more points displacement, higher points velocity, less head displacement, and higher action unit count. These features are consistent with components of the hypothesized startle response, including blinking, hunching the shoulders, grimacing, and baring the teeth. The negative weight for head displacement was somewhat surprising, but we think this observation may be related to subjects freezing in response to being startled. 

In the pain task, higher expressiveness was associated with higher action unit count, higher action unit intensity, and less points velocity. These features are consistent with components of the hypothesized pain response, including grimacing, frowning, wincing, and eye closure. Although the existing literature hypothesizes that body motion increases in response to pain, we found that points velocity has a negative weight. However, we think this finding may be related to increased muscle tension and/or the nature of this specific pain elicitation task (e.g., decreased velocity may be related to the \textit{regulation} of pain in particular).

Finally, in the disgust task, higher expressiveness was associated with higher action unit count, higher action unit intensity, higher points displacement, and higher head displacement. These features are consistent with components of the hypothesized disgust response, including furrowed brows, eye closure, nose wrinkling, upper lip retraction, upward movement of the lower lip and chin, and drawing the corners of the mouth down and back. We believe that the observed head displacement weight may be related to subjects recoiling from the source of the unpleasant odor, which would produce changes in head scale and translation.



\section{Conclusion}

In this paper, we define expressiveness as the extent to which an individual shows his or her feelings, thoughts, or reactions in a given moment. Following this definition, we present a dataset that can be used to model or analyze expressiveness in different emotional contexts using human labels of attributes relevant to visual expressiveness. We propose and test a series of deep learning and statistical models to predict expressiveness from visual data; we also use the latter to understand the relationship between intepretable visual features derived from OpenFace and expressiveness. We find that training models for specific emotional contexts results in better predictive performance that training across contexts. We also find support for our hypothesis that expressiveness is associated with unique features in each context, although several features are also important across all contexts (e.g., the amount and intensity of facial movement). Future work would benefit from attending to the similarities and differences in signals of expressiveness across emotional contexts to construct a more robust predictive model.


\bibliography{ref}
\bibliographystyle{aaai}

\newpage
\appendix
\section{Appendix}

\subsubsection{Amazon Mechanical Turk Questions}
Four questions were proposed to capture aspects of expressiveness:
\begin{enumerate}
    \item How strong is the emotional response of the person in this video to [the stimulus] compared to how strongly a typical person would respond?
    \item How much of any emotion does the person show in this video clip?
    \item How much does the person move any part of their body/head/face in this video clip?
    \item How much does any part of the person's face become or stay tense in this video clip?
\end{enumerate}

\subsubsection{Amazon Mechanical Turk Ratings}
For the first question, the Likert scale was anchored for raters as follows:
\begin{itemize}
    \item 0 - No emotional response / Nothing to respond to
    \item 1 - Weak response
    \item 2 - Typical strength response
    \item 3 - Strong response
    \item 4 - Extreme response
\end{itemize}
For the remaining questions, the Likert scale was anchored:
\begin{itemize}
    \item 0 - A little / None
    \item 1
    \item 2 - Some
    \item 3
    \item 4 - A lot
\end{itemize}

\subsubsection{Video Segmentation} For each task, the following segments were sampled from each full subject/task video combination. Timestamps are in SS (seconds) format. The notation -SS refers to a timestamp SS seconds from the end of the video. Frames do not overlap between segments (that is, the last frame of a segments ending at 03 is the frame prior to the first frame of a segment starting at 03).

\begin{itemize}
    \item Sadness: [00, 03], [03, 06], [30, 33], [33, 36], [--12, --09], [--09, --06], [--06, --03], [--03, --00]
    \item Startle: [03, 06], [06, 09], [09, 12], [12, 15], [15, 18]
    \item Fear: [00, 03], [03, 06], [06, 09], [09, 12], [12, 15], [15, 18], [18, 21]
    \item Pain: [00, 03], [03, 06], [06, 09], [--12, --09], [--09, --06], [--06, --03], [--03, --00]
    \item Disgust: [03, 06], [06, 09], [09, 12], [12, 15]
\end{itemize}

\subsubsection{Pilot Studies on Human Rating Reliability}
To determine which tasks and questions could be annotated with adequate inter-rater reliability, we conducted a pilot study with 3 crowdworkers rating the video clips from 5 subjects on 4 questions. The results of this study are provided in Table~\ref{tab:pilot-n3-icc}. The ICC scores for the sadness and fear tasks looked poor overall, and these tasks were excluded. The ICC scores looked good for the disgust task, and we thought that increasing the number of raters to 6 might increase the reliability of the startle and pain tasks to adequate levels. The results of a follow-up study with 6 crowdworkers are provided in Table~\ref{tab:pilot-n6-icc}. The ICC scores indicate that the first three questions could be annotated with adequate reliability, but the fourth question had poor reliability and was excluded. As such, the final study included 6 raters of the startle, pain, and disgust tasks with the first three questions only.

\begin{table}[h]
    \begin{center}
    \begin{tabular}{lccc}
        \toprule
        Task & Question & ICC & 95\% CI \\
        \midrule
        Sadness & 1 & 0.419 & [--0.014, 0.687] \\
        Sadness & 2 & 0.517 & [0.156, 0.739] \\
        Sadness & 3 & 0.311 & [--0.203, 0.628] \\
        Sadness & 4 & 0.562 & [0.235, 0.764] \\
        \midrule
        Startle & 1 & 0.632 & [0.290, 0.825] \\
        Startle & 2 & 0.616 & [0.248, 0.821] \\
        Startle & 3 & 0.749 & [0.515, 0.861] \\
        Startle & 4 & 0.280 & [--0.391, 0.658] \\
        \midrule
        Fear & 1 & 0.197 & [--0.403, 0.567] \\
        Fear & 2 & 0.391 & [--0.168, 0.639] \\
        Fear & 3 & 0.368 & [--0.103, 0.659] \\
        Fear & 4 & 0.086 & [--0.596, 0.507] \\
        \midrule
        Pain & 1 & 0.652 & [0.392, 0.812] \\
        Pain & 2 & 0.534 & [0.187, 0.749] \\
        Pain & 3 & 0.504 & [0.135, 0.733] \\
        Pain & 4 & --1.010 & [--2.509, --0.084] \\
        \midrule
        Disgust & 1 & 0.879 & [0.747, 0.948] \\
        Disgust & 2 & 0.856 & [0.699, 0.938] \\
        Disgust & 3 & 0.837 & [0.660, 0.930] \\
        Disgust & 4 & 0.797 & [0.568, 0.915] \\
        \bottomrule
    \end{tabular}
    \end{center}
\caption{\label{tab:pilot-n3-icc}Intraclass correlation (ICC) among Amazon Turk raters ($n=3$ raters per question) in 5-subject pilot studies.}
\end{table}

\begin{table}[h]
    \begin{center}
    \begin{tabular}{lccc}
        \toprule
        Task & Question & ICC & 95\% CI \\
        \midrule
        Startle & 1 & 0.783 & [0.619, 0.892] \\
        Startle & 2 & 0.777 & [0.605, 0.891] \\
        Startle & 3 & 0.879 & [0.788, 0.940] \\
        Startle & 4 & 0.103 & [-0.574, 0.553] \\
        \midrule
        Pain & 1 & 0.774 & [0.634, 0.872] \\
        Pain & 2 & 0.765 & [0.620, 0.867] \\
        Pain & 3 & 0.763 & [0.615, 0.868] \\
        Pain & 4 & -0.059 & [-0.711, 0.403] \\
        \bottomrule
    \end{tabular}
    \end{center}
\caption{\label{tab:pilot-n6-icc}Intraclass correlation (ICC) among Amazon Turk raters ($n=6$ raters per question) in 5-subject pilot studies.}
\end{table}

\end{document}